\def\year{2020}\relax
\documentclass[letterpaper]{article} 
\usepackage{aaai20}  
\usepackage{times}  
\usepackage{helvet} 
\usepackage{courier}  
\usepackage[hyphens]{url}  
\usepackage{graphicx} 
\usepackage{graphicx}  
\usepackage{enumitem}
\usepackage{subfigure}
\usepackage{amsmath}
\usepackage{multirow}
\usepackage{booktabs}
\usepackage{amssymb}

\DeclareMathOperator*{\argmax}{argmax}

\title{Relation Network for Person Re-identification}
\author{Hyunjong Park, Bumsub Ham\thanks{Corresponding author} \\ 
School of Electrical and Electronic Engineering, Yonsei University \\ 
hyunpark@yonsei.ac.kr, bumsub.ham@yonsei.ac.kr \\ 
}
 
\begin{document}

\maketitle

\begin{abstract}
Person re-identification~(reID) aims at retrieving an image of the person of interest from a set of images typically captured by multiple cameras. Recent reID methods have shown that exploiting local features describing body parts, together with a global feature of a person image itself, gives robust feature representations, even in the case of missing body parts. However, using the individual part-level features directly, without considering relations between body parts, confuses differentiating identities of different persons having similar attributes in corresponding parts. To address this issue, we propose a new relation network for person reID that considers relations between individual body parts and the rest of them. Our model makes a single part-level feature incorporate partial information of other body parts as well, supporting it to be more discriminative. We also introduce a global contrastive pooling~(GCP) method to obtain a global feature of a person image. We propose to use contrastive features for GCP to complement conventional max and averaging pooling techniques. We show that our model outperforms the state of the art on the Market1501, DukeMTMC-reID and CUHK03 datasets, demonstrating the effectiveness of our approach on discriminative person representations. 
\end{abstract}

\section{Introduction}
Person re-identification~(reID) is one of fundamental tasks in computer vision, with the purpose of retrieving a particular person from a set of pedestrian images, captured by multiple cameras. It has been getting a lot of attention in recent years, due to the wide range of applications including pedestrian detection~\cite{zheng2017person} and multi-person tracking~\cite{tang2017multiple}. This problem is very challenging, since  pedestrians have different attributes~(\emph{e.g.},~clothing, gender, hair), and the pictures of them are taken under different conditions, such as illumination, occlusion, background clutter, and camera types. Remarkable advances in convolutional neural networks~(CNNs) over the last decade allow to obtain person representations~\cite{lin2017consistent,lin2017improving,ge2018fd} robust to these factors of variations, especially for human pose, and they also enables learning metrics~\cite{zhang2016learning,chen2017beyond} for computing the similarities of person features. 

Person reID methods using CNNs typically focus on extracting a global feature of a person image~\cite{lin2017consistent,lin2017improving,ge2018fd,zhang2016learning,chen2017beyond} to obtain a compact descriptor for an efficient retrieval. This, however, gives a limited representation, as the global feature may not account for intra-class variations~(e.g., human pose, occlusion, background clutter). To address this problem, part-based methods~\cite{zhao2017spindle,su2017pose,zheng2017pose,li2018harmonious,liu2017hydraplus,zhao2017deeply,sun2018beyond,fu2018horizontal} have been proposed. They extract local features from body parts~(\emph{e.g.},~arms, legs, torso), often together with the global feature of a person image itself, and aggregate them for an effective person reID. To leverage body parts, these approaches extract pose maps from off-the-shelf pose estimators~\cite{zhao2017spindle,su2017pose,zheng2017pose}, compute attention maps to consider discriminative regions of interest~\cite{li2018harmonious,liu2017hydraplus,zhao2017deeply}, or slice person images into horizontal grids~\cite{sun2018beyond,fu2018horizontal}. Part-level features provide better person representations than a global one, but aggregating the individual local features~\emph{e.g.},~by concatenating them without considering relations between body parts, is limited to represent an identity of a person discriminatively. In particular, this does not differentiate the identities of different persons that have similar attributes in corresponding parts between images, since part-based methods compute the similarity of corresponding part-level features independently.

In this paper, we propose to make each part-level feature incorporate information of other body parts to obtain discriminative person representations for an effective person reID. To this end, we introduce a new relation module exploiting one-vs.-rest relations of body parts. It accounts for the relations between individual body parts and the rest of them, so that each part-level feature contains information of the corresponding part itself and other body parts, supporting it to be more discriminative. As will be seen in our experiments, considering the relation between body parts provides better part-level features with a clear advantage over current part-based methods. We have observed that 1)~directly using both global average and max pooling techniques~(GAP and GMP) to obtain a global feature of a person image does not provide a performance gain, and 2)~GMP gives better results than GAP. Based on this, we also present a global contrastive pooling~(GCP) method to obtain better feature representations based on GMP, which adaptively aggregates GAP and GMP results of the entire part-level features. Specifically, it uses the discrepancy between the pooling results, and distill the complementary information to max pooled features in a residual manner. Experimental results on standard benchmarks, including the Market1501~\cite{zheng2015scalable}, DukeMTMC-reID~\cite{ristani2016performance}, and CUHK03~\cite{li2014deepreid}, demonstrate the advantage of our approach for person reID. To encourage comparison and future work, our code and models are available online:~\url{https://cvlab-yonsei.github.io/projects/RRID/}.

The main contributions of this paper can be summarized as follows:~1) We introduce a relation network for part-based person reID to obtain discriminative local features.~2) We propose a new pooling method exploiting contrastive features, GCP, to extract a global feature of a person image. 3) We achieve a new state of the art, outperforming other part-based reID methods by a large margin.

\begin{figure*}[t]
\centering
	\includegraphics[width=1.4\columnwidth]{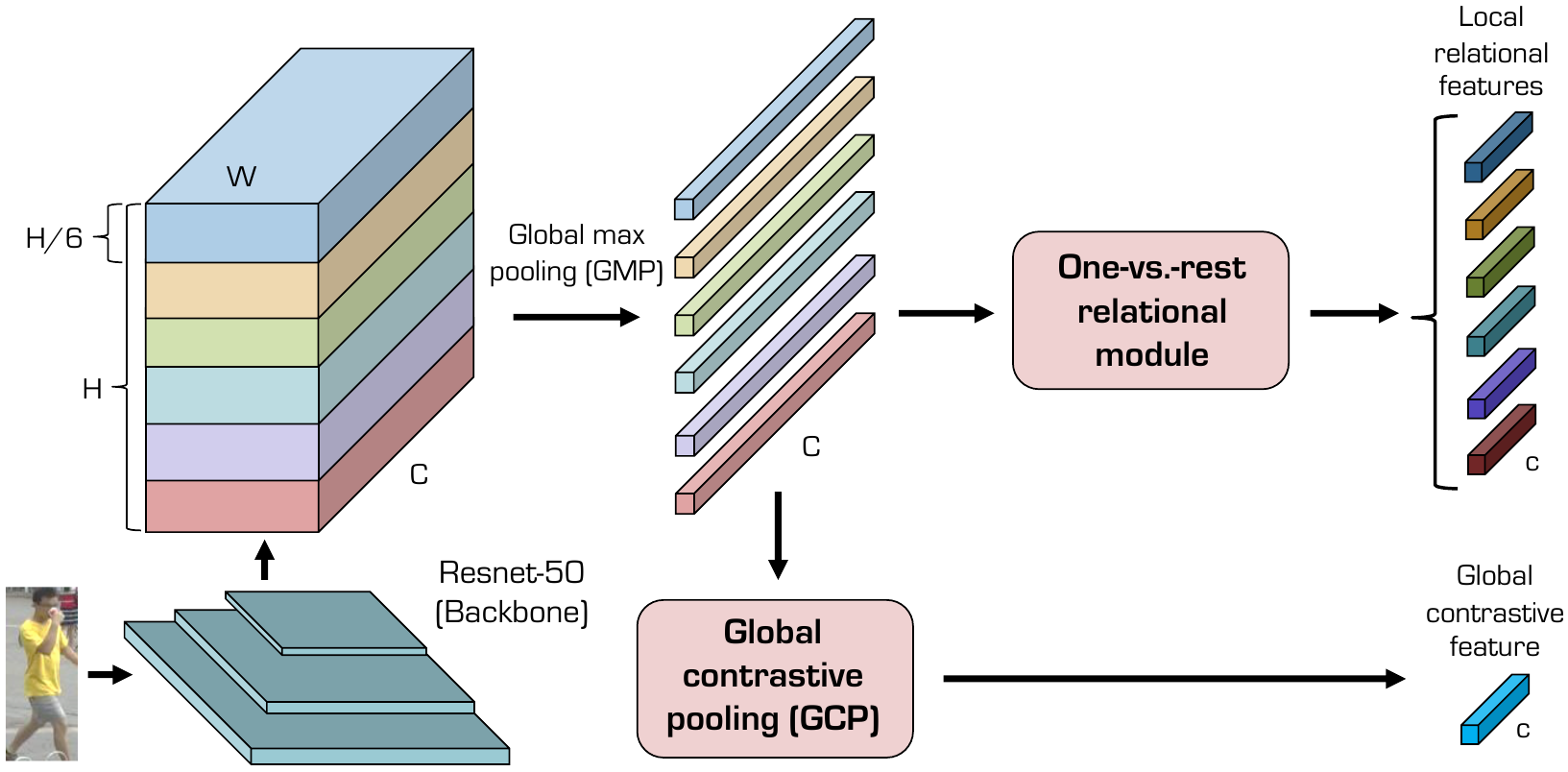}
    \caption{Overview of our framework. The proposed reID model mainly consists of three parts: We first extract part-level features by applying GMP to individual horizontal slices of the feature map from the backbone network. We then input the local features into separate modules, a one-vs.-rest relation module and GCP, that give local relational and global contrastive features, respectively.}
    \label{fig:overview}
\end{figure*}

\section{Related Work}

 \paragraph{Person reID.}

Several person reID methods based on CNNs have recently been proposed. They typically formulate the reID task as a multi-class classification problem~\cite{zheng2018discriminatively}, where person images of the same identity belong to the same category. A classification loss encourages the images of the same identity to be embedded nearby in feature space. Other reID methods additionally use person images of different identities for training, and enforces the feature distance of person images with the same identity to be smaller than that with different identities by a ranking loss. There are many attempts to obtain discriminative feature representations,~\emph{e.g.},~leveraging generative adversarial networks~(GANs) to distill identity-related features~\cite{ge2018fd}, using attributes to offer complementary information~\cite{lin2017improving}, or exploiting body parts to extract diverse person features~\cite{zhao2017spindle,zheng2017pose,su2017pose,liu2017hydraplus,li2018harmonious,yao2019deep,zhao2017deeply,sun2018beyond,fu2018horizontal,wang2018learning}.

Part-based methods enhance the discriminative capabilities of various body parts. We classify them into three categories: The first approach uses a pose estimator~(or a landmark detector) to extract a pose map~\cite{zhao2017spindle,zheng2017pose,su2017pose}. This requires an additional data with landmark annotations to train a pose estimator, and the retrieval accuracy of reID largely depends on the performance of the estimator. The second approach leverages body parts implicitly using an attention map~\cite{li2018harmonious,liu2017hydraplus,zhao2017deeply,yao2019deep}, which can be achieved without auxiliary supervisory signals~(\emph{i.e.},~pose annotations). It provides a feature representation robust to background clutter, focusing on the regions of interest, but the attended regions may not contain discriminative body parts. The third approach also exploits body parts implicitly dividing person images into horizontal grids of multiple scales~\cite{sun2018beyond,fu2018horizontal,wang2018learning}. It assumes that person pictures, localized by off-the-shelf object detectors~\cite{felzenszwalb2008discriminatively}, generally have the same body parts for particular grids~(\emph{e.g.},~legs on the lower parts of person images). This is, however, problematic when the detectors do not localize the persons tightly. Our method belongs to the third category. In contrast to other methods, we aggregate local features while considering relations between body parts, rather than exploiting them directly. Furthermore, we introduce a GCP method to obtain a global feature of a person image, providing discriminative person representations.

  \paragraph{Relation network.}
Exploiting relational reasoning~\cite{santoro2017simple,baradel2018object,sun2018actor,sung2018learning} is important for many tasks requiring the capacity to reason about dependencies between different entities~(\emph{e.g.},~objects, actors, scene elements). Many works have been proposed to support relation-centric computation including interaction networks~\cite{battaglia2016interaction} and gated graph sequence networks~\cite{li2016gated}. The relation network~\cite{santoro2017simple} is a representative method that has been successfully adapted to computer vision problems, including visual question answering~\cite{santoro2017simple}, object detection~\cite{baradel2018object}, action recognition~\cite{sun2018actor}, and few-shot learning~\cite{sung2018learning}. The basic idea behind the relation network is to consider all pairs of entities and to integrate all these relations \emph{e.g.},~to answer the question~\cite{santoro2017simple} or to localize the objects of interest~\cite{baradel2018object}. Motivated by this work, we leverage relations of body parts to obtain better person representations for part-based person reID. Differently, we exploit the relations between individual body parts and the rest of them, rather than considering all pairs of the parts. This encourages each part-level feature to incorporate information of other body parts as well, making it more discriminative, while retaining compact feature representations for person reID.

\section{Our Approach}
  
  We show in Fig.~\ref{fig:overview} an overview of our framework. We extract a feature map of size~$H \times W \times C$ from a person image, where $H$, $W$, $C$ are height, width, and the number of channels, respectively. The resulting feature map is divided equally into six horizontal grids. We then apply GMP to each feature map, and obtain part-level features of size~$1 \times 1 \times C$. We feed these features through two modules in order to extract novel local and global person representations: One-vs.-rest relation module and GCP. The first module makes each part-level feature more discriminative by considering the relations between individual body parts and the rest of them, and outputs local relational features of size~$1 \times 1 \times c$~where $c<C$. The second module provides a global contrastive feature of size~$1 \times 1 \times c$ representing the person image itself. We concatenate global contrastive and local relational features along the channel dimension, and use the feature of size~$1 \times 1 \times 7c$ as a person representation for reID. We train our model end-to-end using cross-entropy and triplet losses, with triplets of anchor, positive and negative person images, where the anchor image has the same identity as a positive one while having a different identity from a negative one. At test time, we extract features of person images, and compute the Euclidean distance between them to determine the identities of persons.

    \begin{figure*}
      \centering
      \subfigure[One-vs.-rest relation module]{\includegraphics[width=0.85\columnwidth]{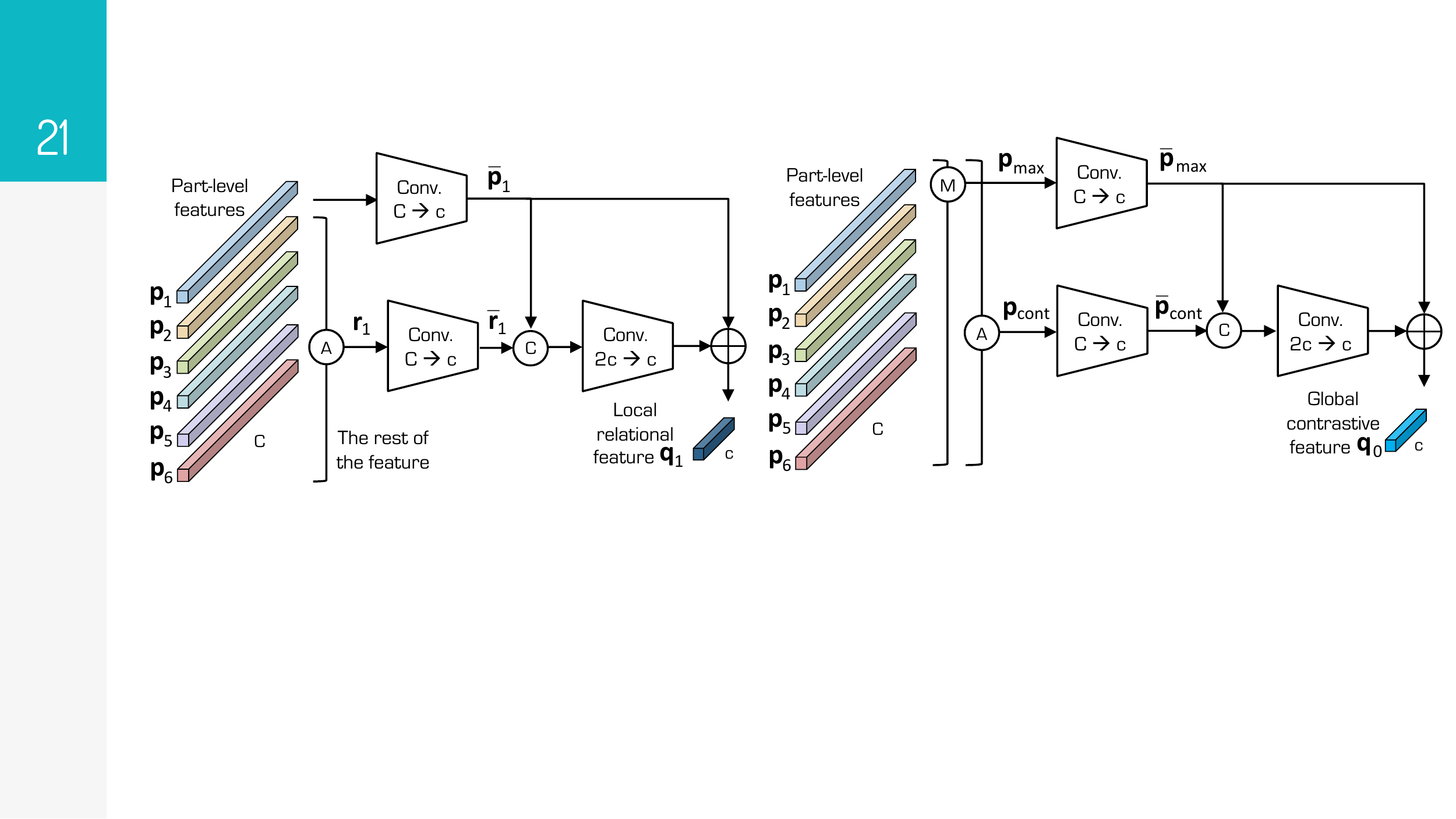}}
      \subfigure[GCP]{\includegraphics[width=.52\linewidth]{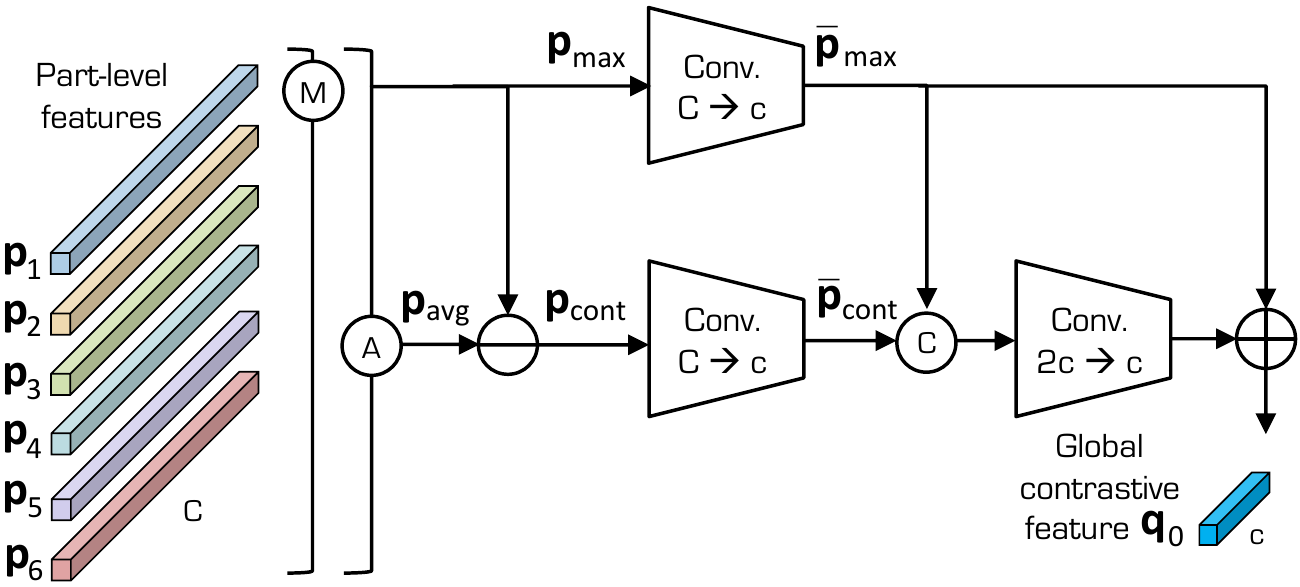}}
      \caption{Illustration of (a) a one-vs.rest relational module and (b) GCP. The relation module gives a local relational feature~$\Vec{q}_i$ for each~${\Vec{p}}_i$. Here, we show a process of extracting the feature~$\Vec{q}_1$. Other relational local features are similarly computed. The GCP outputs a global contrastive feature~$\Vec{q}_0$ considering all part-level features simultaneously. We do not share weight parameters of convolutional layers for all part-level features. See text for details.}
      \label{fig:Details}

    \end{figure*} 
    
    \begin{figure*}
      \centering
      \subfigure[GAP]{\includegraphics[width=.32\columnwidth]{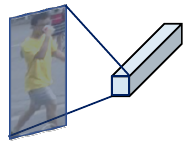}}
      \subfigure[GMP]{\includegraphics[width=.32\columnwidth]{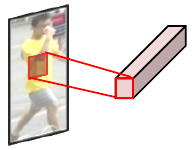}}
      \subfigure[GAP+GMP]{\includegraphics[width=.5\columnwidth]{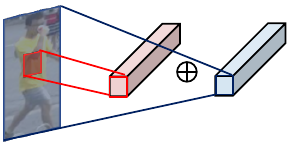}}
      \subfigure[GCP]{\includegraphics[width=.56\columnwidth]{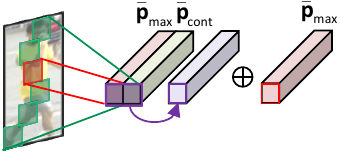}}
      \caption{Illustration of various pooling methods: (a) GAP; (b) GMP; (c) GAP + GMP; (d) GCP. }
      \label{fig:Pooling_methods}

    \end{figure*}   
      
  \subsection{Relation networks for part-based reID}\label{sec:main part}
  
    \paragraph{Part-level features.}
    
We exploit a ResNet-50~\cite{he2016deep} trained for ImageNet classification~\cite{deng2009imagenet} as our backbone network to extract an initial feature map from an input person image. Specifically, following the work of~\cite{sun2018beyond}, we remove the GAP and fully connected layers from the ResNet-50 architecture, and set stride of the last convolutional layer to 1. Similar to other part-based reID methods~\cite{sun2018beyond,fu2018horizontal}, we split the initial feature map into multiple horizontal grids of size~$H/6 \times W \times C$. We apply GMP to each of them, and obtain part-level features of size~$1 \times 1 \times C$.

    \paragraph{One-vs.-rest relational module.}
    
Extracting part-level features from the horizontal grids allows to leverage body parts implicitly for diverse person representations. Existing reID methods~\cite{sun2018beyond,fu2018horizontal,wang2018learning} use these local features independently for person retrieval. They concatenate all local features in a particular order, considering rough geometric correspondences between person images. Although this gives a structural person representation robust to geometric variations and occlusion, the local features cover small parts of an image only, and more importantly they do not account for the relations between body parts. That is, individual parts are isolated, and do not communicate with other ones, which distracts computing the similarity between different persons with similar attributes in corresponding parts. To alleviate this problem, we propose to leverage the relations between body parts for person representations. Specifically, we introduce a new relation network~(Fig.~\ref{fig:Details}(a)) that exploits one-vs-rest relation of body parts, making it possible for each part-level feature to contain information of the corresponding part itself and other body parts.

Concretely, we denote by $\mathbf{p}_i~(i=1,\dots,6)$ each part-level feature of size~$1\times 1 \times C$. We apply an average pooling to all part-level features, except the one of the particular part~$\Vec{p}_i$, aggregating the information from other body parts as follows:~$\Vec{r}_i=\frac{1}{5} \sum_{j\neq i} \Vec{p}_j$. We then add a $1\times 1$ convolutional layer separately for each~$\mathbf{p}_i$ and $\mathbf{r}_i$, giving feature maps~$\bar{\Vec{p}}_i$ and $\bar{\Vec{r}}_i$ of size $1 \times 1 \times c$, respectively. The relation network concatenates the features~$\bar{\Vec{p}}_i$~and~$\bar{\Vec{r}}_i$, and outputs a local relational feature~$\Vec{q}_i$ for each~${\Vec{p}}_i$. We depict in Fig.~\ref{fig:Details}(a) an example of extracting the local relational feature~$\Vec{q}_1$. Here, we assume that the feature~$\Vec{q}_i$ contains information of the original one~$\bar{\Vec{p}}_i$ itself and other body parts. We thus use a skip-connection~\cite{he2016deep} to transfer the relational information of $\bar{\Vec{p}}_i$ and $\bar{\Vec{r}}_i$ to $\bar{\Vec{p}}_i$, as follows:
      \begin{equation}\label{eq:local}
			\Vec{q}_i = \bar{\Vec{p}}_i + R_\mathrm{p}(T(\bar{\Vec{p}}_i, \bar{\Vec{r}}_i)),  \hspace{1.5em}  (i = 1,\dots,6),
      \end{equation}
where $R_\mathrm{p}$ is a sub-network consisting of a $1\times 1$~convolution, batch normalization~\cite{ioffe2015batch}, and ReLU~\cite{krizhevsky2012imagenet} layers. We denote by $T$ a concatenation of features. The residual~$R_\mathrm{p}(T(\bar{\Vec{p}}_i, \bar{\Vec{r}}_i))$ supports the part-level feature~$\bar{\Vec{p}}_i$, making it more discriminative and robust to occlusion. We may leverage all pairwise relations between the features~$\bar{\Vec{p}}_i$ similar to~\cite{battaglia2016interaction}, but this requires a large computational cost and increases the dimension of features drastically. In contrast, our one-vs.-rest relation module computes the feature~$\Vec{q}_i$ in linear time, and also retains a compact feature representation.

    \paragraph{GCP.}
    
To represent an entire person image, previous reID methods use GAP~\cite{sun2018beyond}, GMP~\cite{wang2018learning}, or both~\cite{fu2018horizontal}. GAP covers the whole body parts of the person image~(Fig.~\ref{fig:Pooling_methods}(a)), but it is easily distracted by background clutter and occlusion. GMP overcomes this problem by aggregating the feature from the most discriminative part useful for reID while discarding background clutter~(Fig.~\ref{fig:Pooling_methods}(b)). This, however, does not contain information from the whole body parts. A hybrid approach to exploiting both GAP and GMP~\cite{fu2018horizontal} may perform better, but it is also influenced by background clutter~(Fig.~\ref{fig:Pooling_methods}(c)). It has been proven that GMP is more effective than GAP~\cite{fu2018horizontal}, which will be also verified once more in our experiment. Motivated by this, we propose a novel GCP method based on GMP to extract a global feature map from the whole body parts~(Fig.~\ref{fig:Details}(b)). Rather than applying GAP or GMP to the initial feature map from the input person image~\cite{sun2018beyond,fu2018horizontal}, we first perform average and max pooling with all part-level features. We denote by~$\Vec{p}_\mathrm{avg}$ and $\Vec{p}_\mathrm{max}$ resulting feature maps obtained by average and max pooling, respectively. Note that $\Vec{p}_\mathrm{avg}$ and $\Vec{p}_\mathrm{max}$ are robust to background clutter, as we use a GMP method to obtain the initial part-level features~(Fig.~\ref{fig:overview}). That is, we aggregate the features from the most discriminative parts for every horizontal regions. In particular,~$\Vec{p}_\mathrm{max}$ corresponds to the result of GMP with respect to the initial feature map from the backbone network. We then compute a contrastive feature~$\Vec{p}_\mathrm{cont}$ by subtracting $\Vec{p}_{max}$ from $\Vec{p}_{avg}$, namely, the discrepancy between them. It aggregates most discriminative information from individual body parts~(\emph{e.g.},~green boxes in Fig.~\ref{fig:Pooling_methods}(d)) except the one for $\Vec{p}_{max}$~(\emph{e.g.},~the red box in Fig.~\ref{fig:Pooling_methods}(d)). We add bottleneck layers to reduce the number of channels of~$\Vec{p}_\mathrm{cont}$ and $\Vec{p}_\mathrm{max}$ from $C$ to $c$, denoted by $\bar{\Vec{p}}_\mathrm{cont}$ and $\bar{\Vec{p}}_\mathrm{max}$, respectively, and finally transfer the complementary information of the contrastive feature~$\bar{\Vec{p}}_\mathrm{cont}$ to~$\bar{\Vec{p}}_\mathrm{max}$~(Fig.~\ref{fig:Pooling_methods}(d)). Formally, we obtain a global contrastive feature~$\Vec{q}_0$ of the input image as follows:
 \begin{equation}\label{eq:global}
        \Vec{q}_0 = \bar{\Vec{p}}_\mathrm{max} + R_\mathrm{g}(T( \bar{\Vec{p}}_\mathrm{max}, \bar{\Vec{p}}_\mathrm{cont})),
      \end{equation}
where~$R_\mathrm{g}$ is a sub-network that consists of a $1\times 1$ convolutional, batch normalization~\cite{ioffe2015batch}, and ReLU~\cite{krizhevsky2012imagenet} layers. The global feature~$\Vec{q}_0$ is based on $\bar{\Vec{p}}_\mathrm{max}$, and aggregates the complementary information from the contrastive feature~$\bar{\Vec{p}}_\mathrm{cont}$ with reference to~$\bar{\Vec{p}}_\mathrm{max}$. It thus inherits the advantages of GMP such as the robustness to background clutter while covering the whole body parts. We concatenate the global contrastive feature~$\Vec{q}_0$ in~\eqref{eq:global} and local relational ones~$\Vec{q}_i$~$(i=1,\dots,6)$ in~\eqref{eq:local}, and use it as a person representation for reID.

\begin{table*}[t]
\begin{center}
\footnotesize
\label{table:Comparison}
\resizebox{2\columnwidth}{!}{
\begin{tabular}{ p{16.5em} r c c c c c c c c}
\toprule
\multirow{3}*{\parbox{8em}{Methods}} & \multirow{3}*{F-dim.} & \multicolumn{2}{c}{\multirow{2}*{\parbox{5em}{\centering Market1501}}} & \multicolumn{4}{c}{\parbox{5em}{\centering CUHK03}} & \multicolumn{2}{c}{\multirow{2}*{\parbox{7.5em}{\centering DukeMTMC-reID}}}\\
 & & & & \multicolumn{2}{c}{\parbox{6em}{\centering Labeled}} & \multicolumn{2}{c}{\parbox{6em}{\centering Detected}} & & \\
\cmidrule(lr){3-4}  \cmidrule(lr){5-6} \cmidrule(lr){7-8} \cmidrule(lr){9-10}
 & & \parbox{3em}{\centering mAP} & \parbox{3em}{\centering rank-1} & \parbox{3em}{\centering mAP} & \parbox{2.95em}{\centering rank-1} & \parbox{2.8em}{\centering mAP} & \parbox{2.8em}{\centering rank-1} & \parbox{2.8em}{\centering mAP} & \parbox{2.8em}{\centering rank-1}\\
\midrule
SVDNet~\cite{sun2017svdnet} & 2,048 & 62.1 & 82.3 & 37.8 & 40.9 & 37.3 & 41.5 & 56.8 & 76.7 \\
Triplet~\cite{hermans2017defense} & 128 & 69.1 & 84.9 & - & - & - & - & - & - \\
HA-CNN~\cite{li2018harmonious} & 1,024 & 75.7 & 91.2 & 41.0 & 44.4 & 38.6 & 41.7 & 63.8 & 80.5 \\
Deep-Person~\cite{bai2017deep} & 2,048 & 79.5 & 92.3 & - & - & - & - & 64.8 & 80.9 \\
AlignedReID~\cite{zhang2017alignedreid} & 2,048 & 79.1 & 91.8 & - & - & 59.6 & 61.5 & 69.7 & 82.1 \\
PCB~\cite{sun2018beyond} & 1,536 & 77.3 & 92.4 & - & - & 54.2 & 61.3 & 65.3 & 81.9 \\
PCB+RPP~\cite{sun2018beyond} & 1,536 & 81.0 & 93.1 & - & - & 57.5 & 63.7 & 68.5 & 82.9 \\
HPM~\cite{fu2018horizontal} & 3,840 & 82.7 & 94.2 & - & - & 57.5 & 63.9 & 74.3 & 86.6 \\
MGN~\cite{wang2018learning} & 2,048 & 86.9 & \bf{95.7} & 67.4 & 68.0 & 66.0 & 66.8 & \underline{78.4} & 88.7 \\
\midrule
Ours-S & 1,792 & \underline{88.0} & 94.8 & \underline{73.5} & \underline{76.6} & \underline{69.5} & \underline{72.5} & 77.1 & \underline{89.3} \\
Ours-F & 3,840 & \bf{88.9} & \underline{95.2} & \bf{75.6} & \bf{77.9} & \bf{69.6} & \bf{74.4} & \bf{78.6} & \bf{89.7} \\
\bottomrule
\end{tabular}
}
\caption{Quantitative comparison with the state of the art in person reID. We measure mAP(\%) and rank-1 accuracy(\%) on the Market1501~\cite{zheng2015scalable}, CUHK03~\cite{li2014deepreid} and DukeMTMC-reID~\cite{ristani2016performance} datasets. We denote suffixes ``-S" and ``-F" by our models using $\Vec{q}^\mathrm{P6}$ and $T(\Vec{q}^\mathrm{P2}, \Vec{q}^\mathrm{P4}, \Vec{q}^\mathrm{P6})$, respectively.}
\end{center}
\end{table*}

    \begin{figure}[t]
      \centering
      \includegraphics[width=.7\columnwidth]{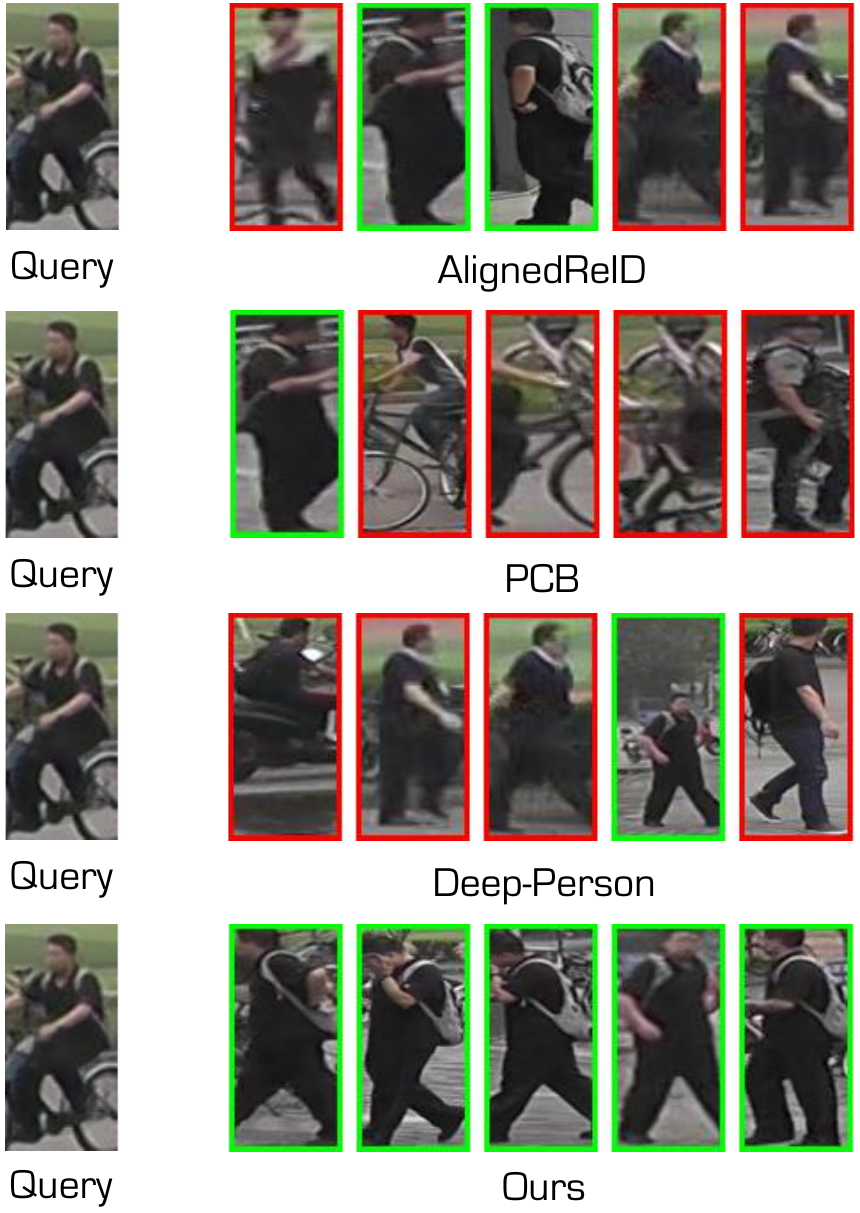}
      \caption{Qualitative comparison of person reID results on the Market1501 dataset~\cite{zheng2015scalable}. We show the top-5 retrieval results~(left:~rank-1, right:~rank-5) for AlignedReID~\cite{zhang2017alignedreid}, Deep-Person~\cite{bai2017deep}, PCB~\cite{sun2018beyond} and Ours-F. Retrieved images with green and red boxes are correct and incorrect results, respectively.}
      \label{fig:Qualitative_comparison}
    \end{figure}
  
  \paragraph{Training loss.}
    
We exploit ground-truth identification labels of person images to learn the person representation. To train our model, we use cross-entropy and triplet losses, balanced by the parameter~$\lambda$ as follows:
       \begin{equation}
        \mathcal{L}=\mathcal{L}_\mathrm{triplet} + \lambda\mathcal{L}_\mathrm{ce},
      \end{equation}
where we denote by $\mathcal{L}_\mathrm{triplet}$ and $\mathcal{L}_\mathrm{ce}$ triplet and cross-entropy losses, respectively. The cross-entropy loss is defines as 
\begin{equation}
	\mathcal{L}_\mathrm{ce} = -\sum_{n=1}^{N} \sum_{i} y^n \log \hat{y}_i^n,
\end{equation}
where we denote by $N$ and $y^n$ the number of images in mini-batch and a ground-truth identification label, respectively. $\hat{y}_i^n$ is a predicted identification label for each feature~$\Vec{q}_i$ in the person representation, defined as 
\begin{equation}
	\hat{y}_i^n = \argmax_{c \in K} \frac{\exp( (\Vec{w}_i^c)^\mathrm{T} \Vec{q}_i )}{\sum_{k=1}^K\exp( (\Vec{w}_i^k)^\mathrm{T} \Vec{q}_i ) }.
\end{equation}
$K$ is the number of identification labels, and $\Vec{w}_i^k$ is classifier for the feature~$\Vec{q}_i$ and the label~$k$. We use a fully connected layer for the classifier. 
To enhance the ranking performance, we use the batch-hard triplet loss~\cite{hermans2017defense}, formulated as follows:
\begin{equation}
\begin{split}
	\mathcal{L}_\mathrm{triplet} = \sum_{k=1}^{N_K} \sum_{m=1}^{N_M} [\alpha + \max_{n=1 \dots M} \Vert \Vec{q}^\mathrm{A}_{k,m} - \Vec{q}^\mathrm{P}_{k,n} \Vert_2 \\
	-  \min_{\substack{l=1\dots K \\ n = 1\dots N \\ l \neq k }} \Vert \Vec{q}^\mathrm{A}_{k,m} - \Vec{q}^\mathrm{N}_{l,n} \Vert_2]_{+}, 
\end{split}
\end{equation}
where $N_K$ is the number of identities in mini-batch, and $N_M$ is the number of images for each identification label in mini-batch~($N = N_K N_M$). $\alpha$ is a margin parameter to control the distances between positive and negative pairs in feature space. we denote by $\Vec{q}^\mathrm{A}_{i,j,}$, $\Vec{q}^\mathrm{P}_{i,j,}$, $\Vec{q}^\mathrm{N}_{i,j,}$ person representations of anchor, positive, and negative images, respectively, where $i$, $j$ correspond to identification and image indexes.

\paragraph{Extension to different numbers of grids.} 

We so far describe our model using global and local features,~\emph{i.e.},~$T(\Vec{q}_0\dots \Vec{q}_6)$, for a person representation, which is denoted by $\Vec{q}^\mathrm{P6}$ hereafter. Without loss of generality, we can use different numbers of horizontal grids for the person representation, to consider various parts of multiple scales~\cite{fu2018horizontal}, such as $\Vec{q}^\mathrm{P2}$ and $\Vec{q}^\mathrm{P4}$ that splits the initial feature map into two and four horizontal regions, respectively. Accordingly, we concatenate the features of $\Vec{q}^\mathrm{P2}$, $\Vec{q}^\mathrm{P4}$ and $\Vec{q}^\mathrm{P6}$, ~\emph{i.e.},~$T(\Vec{q}^\mathrm{P2}, \Vec{q}^\mathrm{P4}, \Vec{q}^\mathrm{P6})$, and use it as a final person representation for an effective reID. Note that $\Vec{q}^\mathrm{P2}$, $\Vec{q}^\mathrm{P4}$, and $\Vec{q}^\mathrm{P6}$ contain different local relational features, and thus have different global contrastive features. Note also that these features share the same backbone network with the same parameters.

\section{Experimental Results} 

  \subsection{Implementation details}
  
    \paragraph{Dataset.}

We test our method on the following datasets and compare its performance with the state of the art. 1)~The Market1501 dataset~\cite{zheng2015scalable} contains 32,668 person images of 1,501 identities captured by six cameras. We use the training/test split provided by~\cite{zheng2015scalable} where it consists of 12,936 images of 751 identities for training and 3,368 query  and 19,732 gallery images of 750 identities for testing. 2)~The CUHK03 dataset~\cite{li2014deepreid} provides 14,097 images of 1,467 identities observed by two cameras, and it offers two types of person images: Manually labeled and detected ones by the DPM method~\cite{felzenszwalb2008discriminatively}. Following the training/test split of~\cite{zhong2017re}, we divide it into 7,365 images of 767 identities for training, and 1,400 query and 5,332 gallery images of 700 identities. 3)~The DukeMTMC-reID~\cite{ristani2016performance} offers 16,522 training images of 702 identities, 2,228 query and 17,661 gallery images of 702 identities.

    \paragraph{Training.}
We resize all images into $384 \times 128$ for training. We set the numbers of feature channels~$C$ to 2,048 and $c$ to 256. This results in 1,792- and 3,840-dimensional features for $\Vec{q}^\mathrm{P6}$ and $T(\Vec{q}^\mathrm{P2}, \Vec{q}^\mathrm{P4}, \Vec{q}^\mathrm{P6})$, respectively. We augment the training datasets with horizontal flipping and random erasing~\cite{zhong2017random}. We use the stochastic gradient descent~(SGD) as the optimizer with momentum of 0.9 and weight decay of 5e-4. We train our model with a batch size~$N$ of 64 for 80 epochs, where we randomly choose 16 identities and sample 4 person images for each identity ($N_K=16$, $N_M=4$). A learning rate initially set to 1e-3 and 1e-2 for the backbone network and other parts, respectively, until 40 epochs is divided by 10 every 20 epochs. We empirically set the weight parameter~$\lambda$ to 2, and fix it to all experiments. All networks are trained end-to-end using \texttt{PyTorch}~\cite{paszke2017automatic}. Training our model takes about six, three and eight hours with two NVIDIA Titan Xps for the Market1501, CUHK03, and DukeMTMC-reID datasets, respectively.

  \subsection{Comparison with the state of the art}
  
    \paragraph{Quantitative results.}

      We compare in Table~\ref{table:Comparison} our models with the state of the arts including part-based person reID methods. We measure mean average precision~(mAP)~(\%) and rank-1 accuracy~(\%) on the Market1501~\cite{zheng2015scalable}, CUHK03~\cite{li2014deepreid}, and DukeMTMC-reID~\cite{ristani2016performance} datasets. We report reID results for a single query for a fair comparison. We denote suffixes ``-S" and ``-F" by our models using $\Vec{q}^\mathrm{P6}$ and $T(\Vec{q}^\mathrm{P2}, \Vec{q}^\mathrm{P4}, \Vec{q}^\mathrm{P6})$, respectively, for final person representations. Table~\ref{table:Comparison} shows that Ours-S outperforms state-of-the-art reID methods in terms of mAP and rank-1 accuracy for all datasets, except MGN~\cite{wang2018learning}. This demonstrates the effectiveness of our one-vs.-rest relation module and GCP. Moreover, Ours-S uses 1,792-dimensional features, allowing to an efficient person retrieval, while providing state-of-the-art results. Ours-F gives the best results again on all datasets in terms of mAP. We achieve mAP of 88.9\% and rank-1 accuracy of 95.2\% for the Market1501, mAP of 69.6\%/75.6\% and rank-1 accuracy of 74.4\%/77.9\% with detected/labeled images for the CUHK03, and mAP of 78.6\% and rank-1 accuracy of 89.7\% for the DukeMTMC-reID. The rank-1 accuracy of Ours-F is slightly lower than that of MGN~\cite{wang2018learning} on the Market1501, but Ours-F outperforms MGN on other datasets by a significant margin. Note that person images in the CUHK03 and DukeMTMC-reID datasets are much more difficult to retrieve, as they are typically with large pose variations, background clutter, occlusion, and confusing attributes.

    \paragraph{Qualitative results.}
    
Figure~\ref{fig:Qualitative_comparison} shows a visual comparison of person retrieval results with the state of the art~\cite{zhang2017alignedreid,bai2017deep,sun2018beyond} on the Market1501 dataset~\cite{zheng2015scalable}. We show the top-5 retrieval results for the query image. The results for all comparisons have been obtained from the official models provided by the authors. We can see that our method retrieves correct person images, and in particular it is robust to attribute variations~(\emph{e.g.},~bicycles) and background clutter~(\emph{e.g.},~grass). Other part-based methods including AlignedReID and PCB try to match local features between images for retrieval. For example, they focus on finding the correspondences for bicycles or grass in the query image, giving many false-positive results. Note that the gallery set contain many images of persons riding a bicycle, but they have different identities from the person in the query.

\begin{table*}[t]
\begin{center}
\footnotesize
\label{table:Ablation study}
\resizebox{2\columnwidth}{!}{
\begin{tabular}{ c c c c c c r c c c c c c c c}
\toprule
\multirow{3}*{GF} & \multirow{3}*{LF} & \multirow{3}*{RM} & \multirow{3}*{Ext.} & \multicolumn{2}{c}{\multirow{3}*{\quad Pooling for}} & \multirow{3}*{F-dim.} & \multicolumn{2}{c}{\multirow{2}*{\parbox{6em}{\centering Market1501}}} & \multicolumn{4}{c}{\parbox{5em}{\centering CUHK03}} & \multicolumn{2}{c}{\multirow{2}*{\parbox{8em}{\centering DukeMTMC-reID}}}\\
 & & & & & & & & & \multicolumn{2}{c}{\parbox{6em}{\centering Labeled}} & \multicolumn{2}{c}{\parbox{6em}{\centering Detected}} & & \\
\cmidrule(lr){8-9}  \cmidrule(lr){10-11}  \cmidrule(lr){12-13} \cmidrule(lr){14-15}

 & & & & GF & LF & & \parbox{2em}{\centering mAP} & \parbox{2.8em}{\centering rank-1} & \parbox{2em}{\centering mAP} & \parbox{2.8em}{\centering rank-1} & \parbox{2em}{\centering mAP} & \parbox{2.8em}{\centering rank-1} & \parbox{3em}{\centering mAP} & \parbox{3em}{\centering rank-1} \\

\midrule
\centering \checkmark & & & & GAP & - & 256 & 74.5 & 88.2 & 57.0 & 60.6 & 54.3 & 58.4 & 62.9 & 79.6 \\
 & \checkmark & & & - & GAP & 1,536 & 79.0 & 92.3 & 65.1 & 67.9 & 62.4 & 65.6 & 70.0 & 84.0 \\
\centering \checkmark & \checkmark & & & GAP & GAP & 1,792 & 82.9 & 92.9 & 68.1 & 71.4 & 63.6 & 66.2 & 73.5 & 85.5 \\
\centering \checkmark & \checkmark & & & GMP & GMP & 1,792 & 83.7 & 93.2 & 70.7 & 73.9 & 64.3 & 66.8 & 74.8 & 86.1 \\
\midrule
\centering \checkmark & \checkmark & \checkmark & & GMP & GMP & 1,792 & 86.7 & 94.2 & 73.3 & 75.8 & 67.6 & 70.2 & 76.3 & 88.3 \\
\centering \checkmark & \checkmark & \checkmark & & GAP & GMP & 1,792 & 85.8 & 94.1 & 72.6 & 75.0 & 67.6 & 69.6 & 75.6 & 88.2 \\
\centering \checkmark & \checkmark & \checkmark & & GAP+GMP & GMP & 1,792 & 86.6 & 94.3 & 72.9 & 75.8 & 68.1 & 70.3 & 76.5 & 88.4 \\
\centering \checkmark & \checkmark & \checkmark & & GCP & GMP & 1,792 & \underline{88.0} & \underline{94.8} & \underline{73.5} & \underline{76.6} & \underline{69.5} & \underline{72.5} & 77.1 & \underline{89.3} \\
\midrule
\centering \checkmark & \checkmark & & \checkmark & GMP & GMP & 3,840 & 86.7 & 94.4 & 72.8 & 74.6 & 67.7 & 69.1 & 76.1 & 87.7 \\
\centering \checkmark & \checkmark & & \checkmark & GAP & GMP & 3,840 & 86.5 & 94.2 & 72.5 & 74.7 & 67.3 & 69.9 & 76.5 & 87.3 \\
\centering \checkmark & \checkmark & & \checkmark & GAP+GMP & GMP & 3,840 & 86.5 & 94.1 & 72.8 & 75.4 & 66.5 & 69.9 & 76.6 & 87.8 \\
\centering \checkmark & \checkmark & & \checkmark & GCP & GMP & 3,840 & 87.3 & 94.5 & 73.0 & 75.9 & 69.0 & 71.6 & \underline{77.4} & 88.3 \\
\midrule
\centering \checkmark & \checkmark & \checkmark & \checkmark & GCP & GMP & 3,840 & \bf{88.9} & \bf{95.2} & \bf{75.6} & \bf{77.9} & \bf{69.6} & \bf{74.4} & \bf{78.6} & \bf{89.7} \\
\bottomrule
\end{tabular}
}
\caption{Quantitative comparison of different network architectures. We measure mAP(\%) and rank-1 accuracy(\%) on the Market1501~\cite{zheng2015scalable}, CUHK03~\cite{li2014deepreid} and DukeMTMC-reID~\cite{ristani2016performance} datasets. Numbers in bold indicate the best performance and underscored ones are the second best. GF:~Global features; LF:~Local features; RM:~One-vs.-rest relational module; Ext.:~An extension to multiple scales.}
\end{center}
\end{table*}    
      
    \subsection{Discussion}
      
      \paragraph{Ablation study.}

We show an ablation analysis on different components of our model in Table~\ref{table:Ablation study}. We compare the performance for several variants of our model in terms of mAP and rank-1 accuracy. From the first three rows, we can see that using both global and local features improves the retrieval performance, which confirms the finding in part-based reID methods. The fourth row shows that GMP gives better results than GAP, as GMP avoids background clutter. The results in the next row demonstrates the effect of local features obtained using our relation module. For example, this gives the performance gains of 3\% and 1\% for mAP and rank-1 accuracy, respectively, for the Market1501 dataset, which is quite significant. From the fifth and eighth rows, we show a comparison of the retrieval performance according to pooling methods for the global feature, and we can see that GCP performs better than GMP, GAP and GMP+GAP in terms of mAP and rank-1 accuracy. For example, compared with GMP+GAP, our GCP improves mAP from 86.6\% to 88.0\% on the Market1501 dataset. The last four rows suggest that exploiting part-level features of multiple scales is important, and using all components performs best.
		
\begin{figure}[t]
  \begin{center}
    \includegraphics[width=0.95\columnwidth]{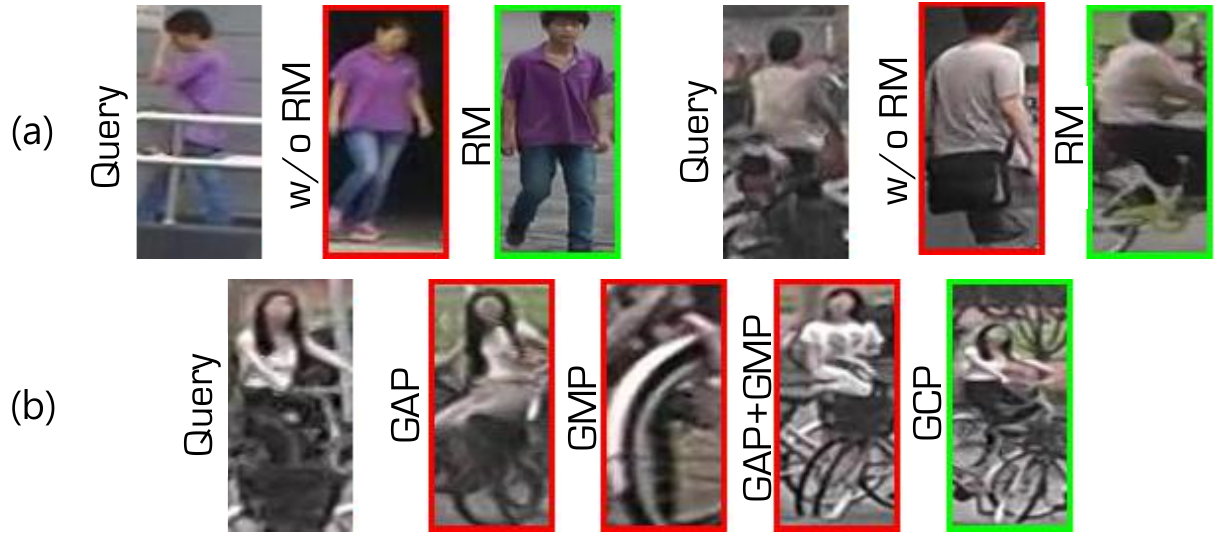}
    \caption{Visual comparison of retrieval results: (a) a relational module and (b) pooling methods. We show top-1 results. The relation module discriminates the same attribute for the person images of different identities. GCP allows to aggregate features from discriminative regions, and provides a person representation robust to background clutter, overcoming the drawbacks of GAP and GMP.}
  \label{fig:qualitative}
  \end{center}
\end{figure}

We show in Fig.~\ref{fig:qualitative}(a) retrieval results with/with a relational module. We can see that person representations obtained from the relation module successfully discriminate the same attribute~(\emph{e.g.}, violet shirts) for the person images of different identities~(\emph{e.g.}, gender), and they are robust to occlusion~(\emph{e.g.}, the person occluded by a bag or a bicycle). We compare in Fig.~\ref{fig:qualitative}(b) retrieval results for different pooling methods. We confirm once again that GAP is not robust to background clutter and GMP sees the most discriminative region~(\emph{e.g.}, bicycle) only rather than the person. We observe that GCP alleviates these problems while maintaining the advantage of GMP.

\begin{figure}[t]
  \begin{center}
    \includegraphics[width=0.8\columnwidth]{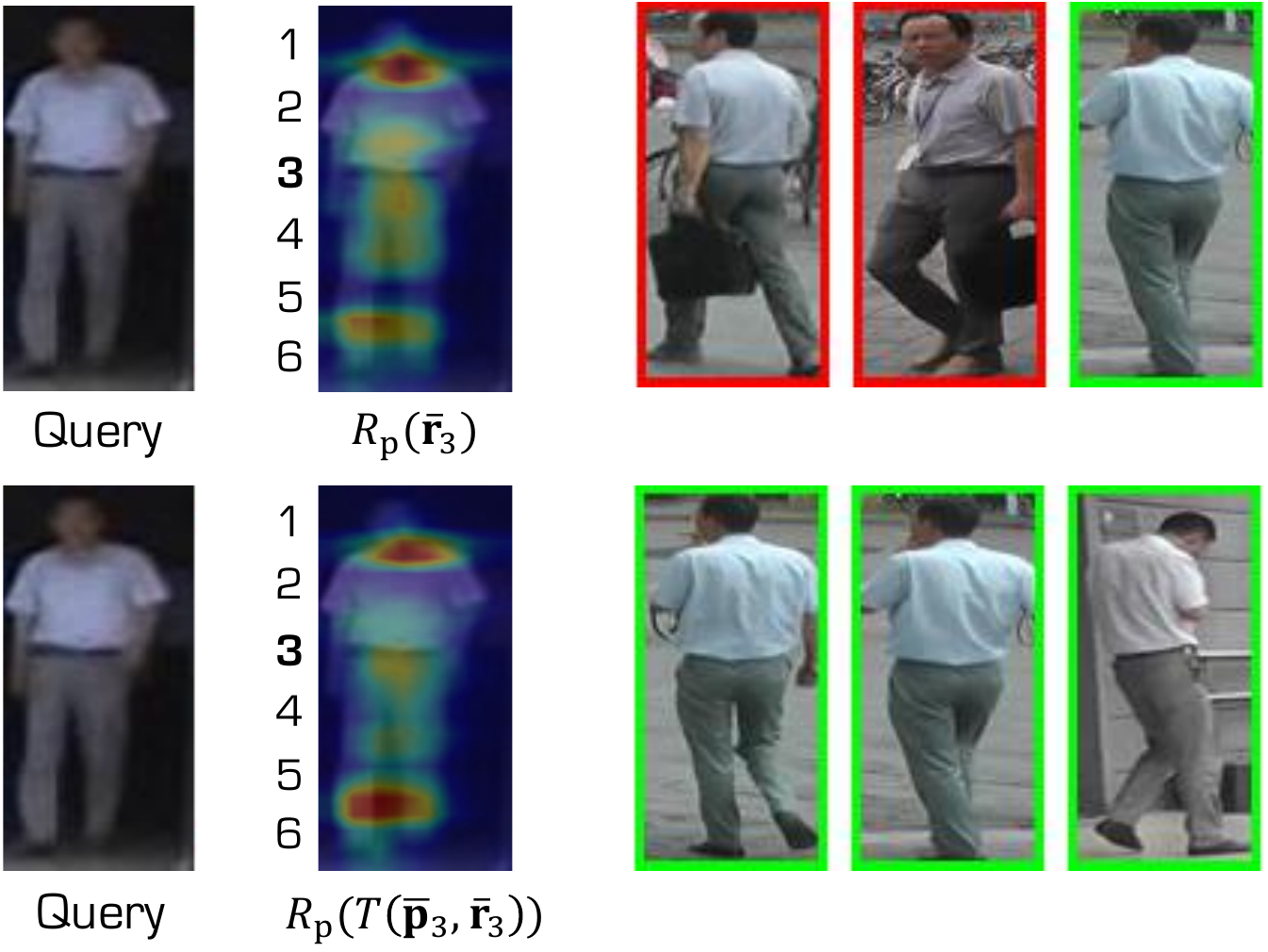}
    \caption{Examples of activation maps for the models using $R_p(\bar{\Vec{r}}_3)$~(top) and $R_p(T(\bar{\Vec{p}}_3, \bar{\Vec{r}}_3))$~(bottom). We also show top-3 retrieval results for each model. We can see that the rest regions~(\emph{e.g.}, the fifth and sixth horizontal grids) are highly activated by the person representation using $R_p(T(\bar{\Vec{p}}_3, \bar{\Vec{r}}_3))$, compared to that using $R_p(\bar{\Vec{r}}_3)$, indicating that our one-vs.-rest relation allows each part-level feature to see the rest regions effectively.}  
  \label{fig:relation_qualitative}
  \end{center}
\end{figure}

	\paragraph{One-vs.-rest relation.}
	We consider relations between each part-level feature $\bar{\Vec{p}}_i$ and its rest feature $\bar{\Vec{r}}_i$. To show how the relation module works, we train a model using the combined rest feature $\bar{\Vec{r}}_i$ instead of $T(\bar{\Vec{p}}_i, \bar{\Vec{r}}_i)$ in~\eqref{eq:local}. In this case, we do not use a multi-scale extension. We visualize activation maps of $R_p(\bar{\Vec{r}}_i)$ and $R_p(T(\bar{\Vec{p}}_i, \bar{\Vec{r}}_i))$ in Fig.~\ref{fig:relation_qualitative}. We observe that $R_p(T(\bar{\Vec{p}}_i, \bar{\Vec{r}}_i))$ focuses more on the regions whose attributes are different from those of $\bar{\Vec{p}}_i$, compared to $R_p(\bar{\Vec{r}}_i)$. This demonstrates that $R_p(T(\bar{\Vec{p}}_i, \bar{\Vec{r}}_i))$ extracts the complementary features of $\bar{\Vec{p}}_i$, that are helpful for person reID but not contained in $\bar{\Vec{p}}_i$, from the rest of the parts. This also verifies that using the feature of $\bar{\Vec{p}}_i$ only is not enough to discriminate the identities of different persons having similar attributes in corresponding parts between images. The mAP/rank-1 for~$\bar{\Vec{r}}_i$ on Market1501 are 84.5/93.4 which are lower than 86.7/94.2 obtained using $T(\bar{\Vec{p}}_i, \bar{\Vec{r}}_i)$ in Table~\ref{table:Ablation study}.

    \paragraph{Performance comparison of local features.}
  
We demonstrate the capabilities of providing discriminative features of the one-vs.-rest relation module. We extract a local feature of size~$1 \times 1 \times 256$ for each horizontal region with and without using the relation module. Given a query image, we then retrieve person images using a single local feature in the person representation of $\Vec{q}^\mathrm{P6}$. We report mAP and rank-1 accuracy for individual local features extracted from different horizontal regions~(1: top, 6: bottom), on the Market1501 dataset~\cite{zheng2015scalable} in Table~\ref{table:Performance_comparison_of_local_features}. From this table, we can observe two things: (1)~The relation module improves mAP and rank-1 accuracy drastically. The improvements in mAP and rank-1 accuracy are 21.7\% and 19.6\%, respectively, on average. The rank-1 accuracy measures the performance of retrieval results for the easiest match, while mAP characterizes the ability to retrieve all person images of the same identity, indicating that the relation module is beneficial especially for retrieving challenging person images. (2)~The local features from the third and last horizontal regions give the best and the worst results, respectively. This suggests that the middle part typically corresponding to the torso of a person provides the most discriminative feature for specifying a particular person, and the bottom part~(\emph{e.g.},~legs or sometimes background due to the incorrect localization of the person detector) gives the least discriminative feature for person reID. 

\begin{table}[t]
\begin{center}
\label{table:Performance_comparison_of_local_features}
\footnotesize
\resizebox{1\columnwidth}{!}{
\begin{tabular}{c c c c c}
\toprule
\multirow{2}*{\parbox{6em}{\centering Horizontal region}} & \multicolumn{2}{c}{\parbox{6em}{\centering w/o RM}} & \multicolumn{2}{c}{\parbox{5em}{\centering RM}} \\
\cmidrule(lr){2-3}  \cmidrule(lr){4-5}
 & \parbox{3em}{\centering mAP} & \parbox{3em}{\centering rank-1} & \parbox{3em}{\centering mAP} & \parbox{3em}{\centering rank-1} \\
\midrule
1 & 29.6 & 49.4 & 58.7 & 82.2 \\
2 & 46.4 & 69.2 & 68.5 & 87.1 \\
3 & \bf{55.3} & \bf{77.2} & \bf{73.5} & \bf{89.1} \\
4 &  \underline{51.8} &  \underline{72.7} &  \underline{71.9} & \underline{87.6} \\
5 & 49.7 & 69.2 & 69.7 & 85.4 \\
6 & 35.5 & 53.4 & 55.7 & 76.9 \\
\bottomrule
\end{tabular}
}
\caption{Quantitative comparison of single local features on the Market1501 dataset~\cite{zheng2015scalable}. RM:~One-vs.-rest relational module.}
\end{center}
\end{table}

    \paragraph{Performance comparison of global features.} 
    
Table~\ref{table:Performance_comparison_of_global_features} compares the reID performance of single global features in terms of mAP and rank-1 accuracy on the Market1501 dataset~\cite{zheng2015scalable}. We use only the 256-dimensional global feature in the person representation of~$\Vec{q}^\mathrm{P6}$ for person retrieval but obtained by different pooling methods. Note that the size of the global feature is much smaller than typical person representations in Table~\ref{table:Comparison}. We can see from Table~\ref{table:Performance_comparison_of_global_features} that GCP gives the best retrieval results in terms of both mAP and rank-1 accuracy, outperforming GAP, GMP, and GAP+GMP by a large margin. Compared with other reID methods in Table~\ref{table:Comparison}, GCP offers a good compromise in terms of the accuracy and the size of features. For example, our global contrastive feature of size~$1\times 1\times 256$ achieves rank-1 accuracy of~93.4\%, which is comparable with 93.1\% for PCB+RPP~\cite{sun2018beyond} using 1,536-dimensional features.

\begin{table}[t]
\begin{center}
\label{table:Performance_comparison_of_global_features}
\footnotesize
\resizebox{1\columnwidth}{!}{
\begin{tabular}{c c c}
\toprule
\multirow{1}*{\parbox{9.5em}{\centering{Methods}}}& \parbox{5.5em}{\centering mAP} & \parbox{5.5em}{\centering rank-1} \\
\midrule
GAP & 81.0 & 91.3 \\
GMP & 81.9 & 92.0 \\
GAP+GMP & \underline{82.5} & \underline{92.6} \\
GCP & \bf{84.6} & \bf{93.4} \\
\bottomrule
\end{tabular}
}
\caption{Quantitative comparison of global features on the Market1501 dataset~\cite{zheng2015scalable}.}
\end{center}
\end{table}  

\section{Conclusion}

We have presented a relation network for person reID considering the relations between individual body parts and the rest of them, making each part-level feature more discriminative. We have also proposed to use contrastive features for a global person representation. We set a new state of the art on person reID, outperforming other reID methods by a significant margin. The ablation analysis clearly demonstrates the effectiveness of each component in our model.

\paragraph{Acknowledgments.}

This research was supported by R\&D program for Advanced Integrated-intelligence for Identification~(AIID) through the National Research Foundation of KOREA~(NRF) funded by Ministry of Science and ICT (NRF-2018M3E3A1057289).

\bibliographystyle{aaai}
\fontsize{9.5pt}{10.5pt} \selectfont
\bibliography{RRID}

\end{document}